\documentclass{article}

 \usepackage[preprint]{neurips_2026}

\usepackage[utf8]{inputenc} % allow utf-8 input
\usepackage[T1]{fontenc}    % use 8-bit T1 fonts
\usepackage{hyperref}       % hyperlinks
\usepackage{url}            % simple URL typesetting
\usepackage{booktabs}       % professional-quality tables
\usepackage{amsfonts}       % blackboard math symbols
\usepackage{nicefrac}       % compact symbols for 1/2, etc.
\usepackage{microtype}      % microtypography
\usepackage{xcolor}         % colors
\usepackage{graphicx}         % colors
\usepackage{amsmath} 

\title{Augmenting Human Evaluation with LLM Judges: How Many Human Reviews Do You Need?}

\author{%
  Jane Paik Kim \\
  Department of Psychiatry and Behavioral Sciences\\
  Stanford University\\
  Stanford, CA 94304 \\
  \texttt{janepkim@stanford.edu} \\
}

\begin{document}

\maketitle

\begin{abstract}
Large language models (LLMs) are increasingly used as automated evaluators of AI systems, including in high-stakes applications.  In this role, LLMs are used to generate judgments about the quality, appropriateness, or even safety of model outputs. This approach is motivated by practical constraints. Expert human ratings are costly and difficult to scale, whereas LLM ratings can be produced quickly at low cost. However, current approaches to deploying LLM evaluators are ad hoc, typically limited to reporting agreement metrics between human and LLM judges as a justification for substitution of human ratings, and lack a formal basis for study design. This paper (1) shifts the role of the LLM judge from substitutive to auxiliary, and (2) formulates the LLM-as-a-judge paradigm as one of augmenting human evaluation through a two-stage sampling design, where LLM evaluations are measured for all observations at the first stage and human ratings are partially observed for a subsample at the second stage.  We propose to use a doubly robust estimator from the missing data literature, which takes advantage of the robustness property against the prediction model, since the missingness model is known by design. Using the asymptotic variance of this estimator, we propose how sample sizes of human and LLM ratings can be determined to achieve a targeted level of power. We also show that a study can be efficiently designed by allocating more human ratings for types of evaluations where the predictability of LLM ratings is not high. To the best of our knowledge, there is very little guidance on how much human oversight should be retained when validating benchmarks.  
\end{abstract}

\section{Introduction}

LLMs are increasingly evaluated on specialized tasks such as clinical summarization, diagnostic interpretation, and patient-facing communication. (\cite{bean2026, croxford2025development, kumar2026}). Evaluating the adequacy of LLM-generated outputs requires domain expertise, as adequacy depends on accuracy, context-sensitive interpretation, and appropriateness. However, expert review is expensive and slow to obtain. To address this bottleneck, researchers have adopted LLM-as-judge approaches, in which LLM outputs are evaluated by other LLMs rather than by human experts (\cite{gu2024}). These approaches scale at negligible cost but introduce a new problem.

A critical problem is that these approaches rely solely on LLM ratings after ad-hoc justification. In practice, this substitution is justified through agreement-based validation. In a common approach, expert ratings are collected on a convenience sample of benchmark instances and compared to LLM-generated ratings using agreement metrics. If agreement meets a chosen threshold, the LLM is treated as a valid replacement, and subsequent evaluation relies on LLM ratings alone (\cite{li2024llms}). A similar pattern appears in automated evaluator pipelines, where expert-labeled test sets are used to validate a trained scoring model by demonstrating agreement with human raters, after which the model is deployed without incorporating expert labels. In both cases, the role of human oversight is solely to gauge the performance of the LLM judge. Once validation is complete, the human ratings are discarded and not used. This strategy may be adequate for low-stakes benchmarking, but in clinical quality monitoring, mental health safety auditing, and regulatory assessment, undetected errors carry meaningful consequences.

There are several limitations to the ad hoc approach of using agreement as the primary justification for replacing human review. High agreement does not establish that LLM judges are rating the same constructs as humans (\cite{chehbouni2025}). Second, agreement is typically heterogeneous across item types, content domains, or evaluation dimensions. \cite{kumar2026} demonstrated this by showing expert-LLM agreement ranged from 0.17 to 0.86 of weighted kappa  across 21 evaluation dimensions of empathic communication. Third, validation sample sizes are not typically justified.

\paragraph{Addressing the gap in rigor.} This paper shifts the dominant framing in LLM evaluation from replacement to augmentation. We propose using LLM-generated ratings as auxiliary data to complement a carefully designed subsample of human evaluations.  We propose framing the LLM-as-a-judge paradigm as a two-stage sampling problem (\cite{zhao1992}), where inexpensive ratings can be measured for the whole sample and expensive ratings can be measured for only a subset of the whole sample.  Under this design, expensive human ratings are partially observed due to cost. To handle incomplete data, one can either use auxiliary data to predict and impute missing values, or weight the observed data by the inverse of the response probability. In typical missing data problems, the validity of these methods depend on the correct specification of the prediction or the response probability model. The doubly robust (DR) estimator is the combination of the two and requires \textit{either} model to be correct for the inference to be valid (\cite{robins1994}). However, in our case, the response probability is dictated by design and is known, and thus the validity of inferences based on the DR estimator is guaranteed.

\paragraph{Our contributions.} A critical direct consequence of this framework is the allowance of prospective designs for LLM-as-a-judge evaluations with formal design components. The form of the asymptotic variance of the DR estimator allows us to determine how many human and LLM samples are needed.  This reframing shifts the emphasis from post-hoc comparison to prospective study design, and provides the sample size formulas and allocation guidelines.  It enables evaluation studies to be designed with the same inferential rigor applied to clinical trials. To the best of our knowledge, there is very little guidance on how much human oversight should be retained when validating benchmarks.

Our primary contributions are (i) to frame the LLM-as-a-judge paradigm as a two-stage design by treating human expert ratings as the primary quantity of inferential interest and LLM-generated ratings as auxiliary measurements that are inexpensive and scalable to obtain and apply missing data methodology, (ii) to provide a sample size calculation for LLM and human ratings given how predictable LLM ratings are for human ratings, and (iii) to provide an efficient allocation strategy to design an LLM-as-a-judge study.

\section{Methods}\label{sec2}

\subsection{Two-stage sampling}

\begin{figure}[htbp] % [htbp] suggests placement: here, top, bottom, page
    \centering
    \includegraphics[width=0.75\textwidth]{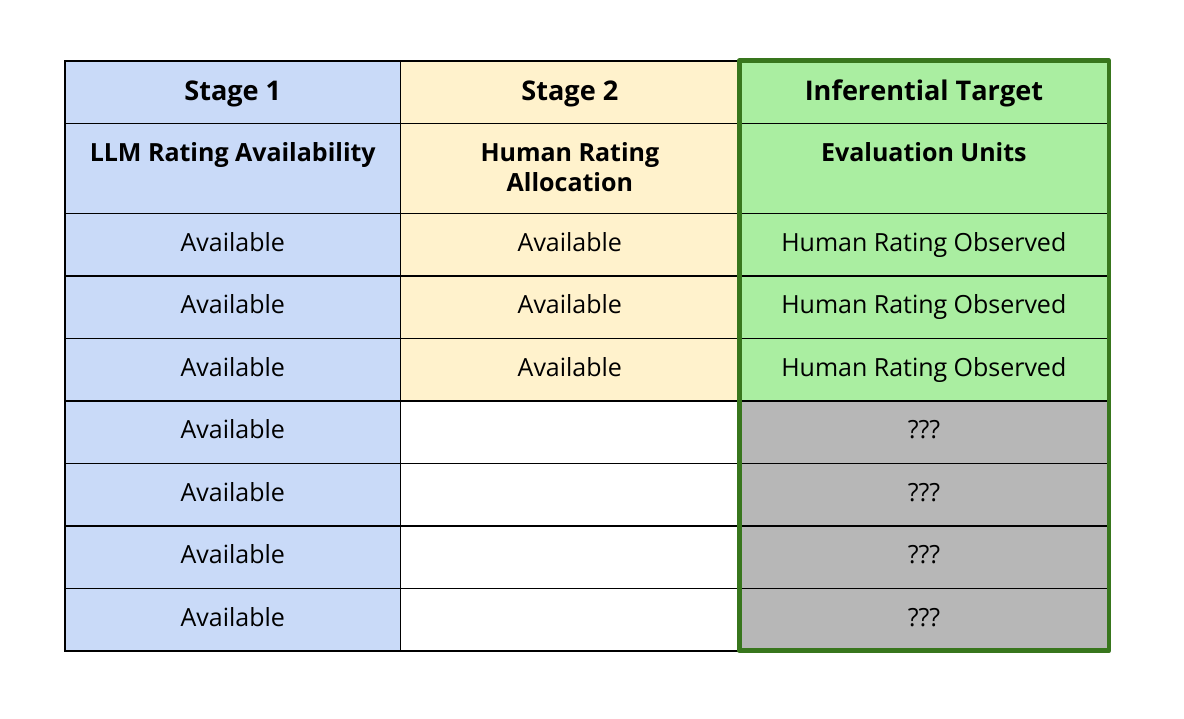}
    \caption{Proposed formulation of a two-stage design: At the first stage, LLM ratings are collected for all evaluation units. At the second stage, human ratings data are collected for a subsample of units, and the remaining units have no observed human rating. The outlined box in green represents the target of inferential interest, the full distribution of human ratings across all units, including incomplete observations.}
    \label{fig:Fig1}
\end{figure}

Our framework first assumes that human ratings are the gold standard of some well defined construct and are of primary inferential interest, and that LLM ratings are available as auxiliary data. We propose a two-stage sampling design, where in the first stage, LLM ratings are available for all evaluative units. In the second stage human ratings are available in only a subset of the first-stage. The role of LLM ratings is to serve as auxiliary data that supplement an incomplete set of human evaluations. This is in contrast with the widely adopted approach of using human ratings as a means to justify the use of LLM ratings but discarded in the main analysis. In a two-stage design, missing data methodology is used to handle incomplete human ratings (Figure~\ref{fig:Fig1}). One distinctive feature of two-stage sampling is that the missingness mechanism is known and determined by design.

\subsection{The estimator}

To motivate the estimator, let us start with some simple intuition. One way to handle missing data is to build a prediction model for human ratings using LLM ratings as a predictor, and fill in the predicted values in place of missing data. This is known as a prediction approach \cite{kim2021}, and its validity depends on the correct model specification of the prediction model. Another way to handle missing data is to use the observed data only but weight the observations by the inverse of the probability of observation or response. This approach is called inverse probability weighting, or propensity score methods, first proposed by \cite{horvitz1952}, and the validity depends on the correct model specification of the response model. The prediction and response models are nuisance models that are not needed if we do not have missing data. The propensity score approach is known to be inefficient (\cite{hahn1998role, heckman1997matching}), but can be made efficient by constructing estimating equations with the subtraction of an unbiased term that renders the resulting estimating function orthogonal to the nuisance model estimating function (\cite{kim2021}). The term that is subtracted involves the predictor of the outcome. We refer to the resulting estimating approach as the doubly robust approach. 

A doubly robust estimator is a weighted average of both estimators but its validity depends on the correctness of either the prediction or response probability model.  As mentioned earlier, the incompleteness of human ratings is a consequence of design so that the probability of observation or response is known and will always be correctly specified. That is, the DR estimator will guarantee that the parameter estimate for the expert human rating population is still valid even if the prediction model is incorrect.

Let $N$ denote the total number of items requiring evaluation. 
Let $\delta_i = 1$ if a human response is observed, 0, otherwise. Let $Y_i$ be the human ratings, and $X_i,$ the predictors of $Y_i,$ including the LLM ratings, and $\pi_i = P(\delta_i =1 | X_i),$ the sampling probabilities. Let $\sum_{i = 1}^N \delta_i = n.$
LLM ratings are observed for all $N$ items, and human ratings are observed only for the $n$ reviewed items. The response probabilities $\pi_i$ are controlled by the investigator. 

Let $U(\theta; X, Y)$ be any estimating function. A simple example is when $U(\theta; X_i, Y_i) = Y_i - \theta.$

Let

\begin{equation}
 W(\theta)  = \sum_{i=1}^N \left[ \frac{\delta_i }{\pi_i} U(\theta; X_i, Y_i)  \right] -
  \sum_{i=1}^N (\frac{\delta_i }{\pi_i} - 1 ) E \left[  U (\theta; X_i, Y_i )  \middle| X_i  \right]
\end{equation}

The solution of $W(\theta) = 0$ gives the DR estimator. 
The underlying population parameter $\theta$ is estimated using a doubly robust estimator (\cite{robins1994, kim2021}). 
When we replace LLM ratings for human ratings, we have $E(Y_i|X_i) = X_i.$ Equation (1) shows that for $\delta = 0,$ the LLM rating will replace the human rating. When $\delta = 1$, a human rating is used but is adjusted by a weighted residual of  human and LLM ratings.

\subsection{Variance of the Doubly Robust Estimator} 

The variance of the doubly robust estimator is shown in (2) and depends on two sources of uncertainty. The first term of the variance represents the variance as if all $N$ values are from human ratings. The second is the penalty for using LLM-ratings in place of human ratings, and depends on two quantities: the residual error, measuring how well the LLM predicts human ratings, and the response or observation probability $\pi_i$.

The variance of the DR estimator is given by the following:

\begin{align}
V \left[ W(\theta) \right] & = V \left[ \frac{1 }{n} \sum_{i=1}^n U(\theta; X_i, Y_i)  \right]  \nonumber \\
& + 
   E \left[ \frac{1 }{n^2} \sum_{i=1}^n (\frac{1}{\pi_i} - 1) \{ U(\theta; X_i, Y_i)- E[U(\theta; X_i, Y_i)\mid X_i ] \}^2  \right]
\end{align}

\noindent This variance formula will be the basis for the sample size calculations.

\section{Main Results}

\subsection{Sample Size Calculations }

Let us begin from the point where the usual sample size calculation is completed, assuming that human ratings are obtainable for all units of assessment (\cite{cohen2013}). Let that sample size be $n^*.$ It is assumed that we need $n^*$ human rating samples to achieve either a desired power or pre-specified level of precision. Consider a simple case of estimating the mean of $Y$ and the variance of $Y$ is $\sigma^2.$ Then the target variance of the sample mean is $\sigma^2 / n^*.$

In practice, obtaining the target sample size $n^*$ of human ratings may not be feasible. Our goal is to find a combination of a human sample of size $n \le n^*,$ an LLM sample of size $N \ge n^*,$ and response probabilities $\pi_i$'s that results in the same target variance  $\sigma^2 / n^*.$  Sometimes, the size of the human sample may be constrained due to feasibility or a given budget. In that case, given a desired $n^*$ and $n$, we can find $N$.  To use the sample size formula, we need some external information on the conditional variance, 

\begin{align*}
E \left[ U(\theta; X_i, Y_i) - E\{ U(\theta; X_i, Y_i)|X_i \} \middle | X_i \right]^2.
\end{align*}

We use the relationship  
\begin{align*}
\frac{ \sigma_e^2}{ \sigma^2}=1-\rho ^2
\end{align*}

where 
\begin{align*}
\sigma^2_e = E[\{ U(\theta; X_i, Y_i) - E[U(\theta; X_i, Y_i)|X_i)] \}^2 | X_i], \\
\sigma^2 = E\left[ U(\theta; X_i, Y_i) - E\{U(\theta; X_i, Y_i) \} \right]^2, 
\end{align*}

and
%$\sigma^2 = E(U - E[U(\theta, X_i, Y_i))])^2,$ and 
\begin{align*}
\rho ^2 = corr(U(\theta; X_i, Y_i), \hat E[U(\theta; X_i, Y_i)|X_i)])^2.
\end{align*}

The conditional variance is related to the correlation between human and LLM ratings. We denote an estimate of the correlation $\rho^2$ by $R^2$. 

For $U(\theta; X_i, Y_i) = Y_i - \theta$ and $\pi_i = \pi$ for all $i$, the variance equation (2) reduces to 

\begin{align}
 \frac{\sigma^2}{n^*}  & \approx  \frac{1}{N}  \left[ \sigma^2 + \frac{1 - \pi}{ \pi} \cdot \sigma^2(1 - \rho ^2) \right].  
\end{align}

Given $(n^*, R^2)$ we can obtain multiple pairs of $(N, \pi),$ or equivalently, $(N, n),$ that satisfy (3).

\begin{figure}[htbp]
\centering
\includegraphics[width=0.8\textwidth]{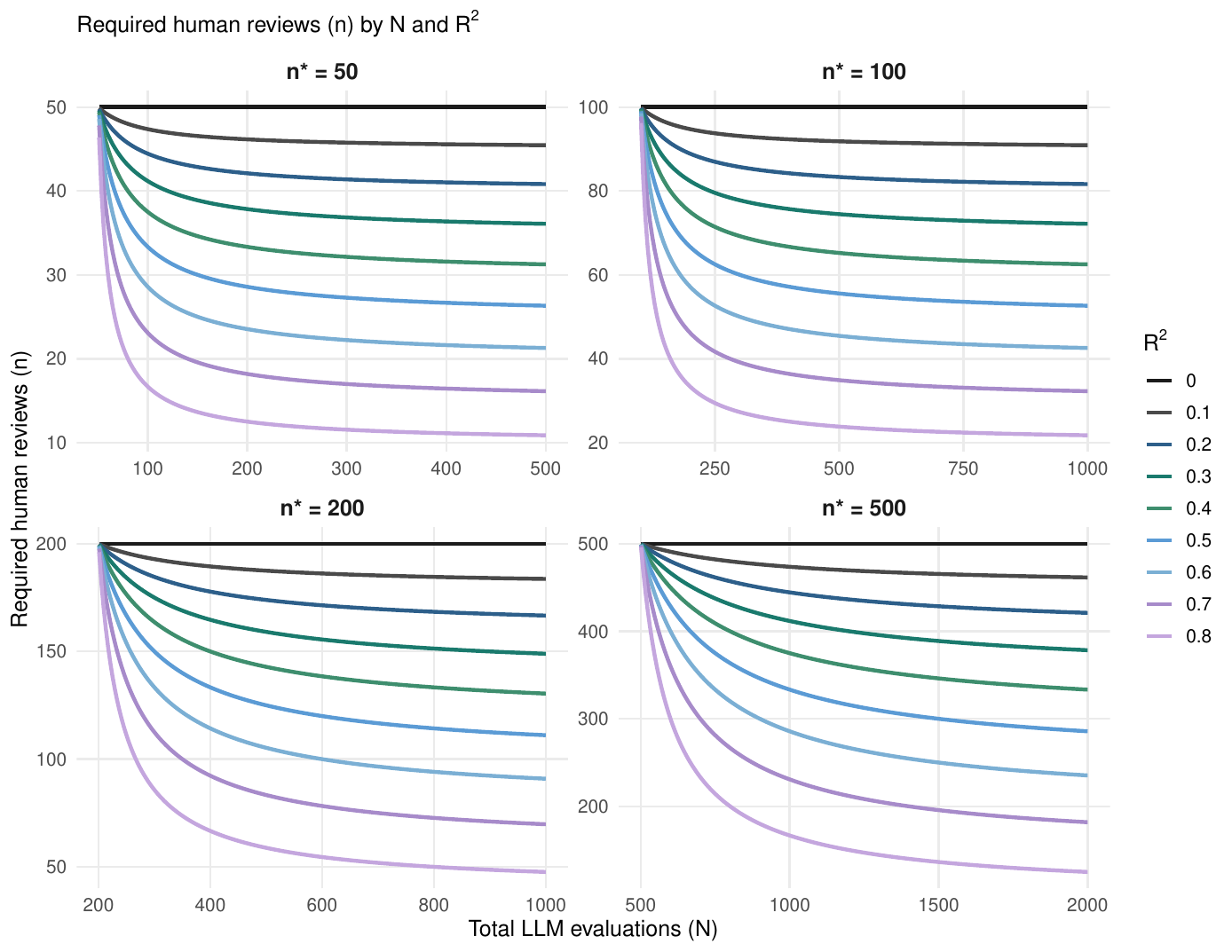}
\caption{Number of human reviews as function of total LLM evaluation, for a given target effective sample size. The total sample is expressed as a multiple of the target. At $R^2=0$, the LLM judge does not contribute and the total target sample size is equivalent to the required number of human samples. }
\label{fig:Fig2}
\end{figure}

Figure~\ref{fig:Fig2} presents the required human sample size, $n$, given the LLM sample size $N,$ stratified by the range of $R^2$ for effective sample sizes $n^* = 50, 100, 200, 500$.  The required number of human reviews decreases in the LLM sample size when fixing $R^2$, as well as decreases in $R^2$ when holding the LLM sample size $N$ constant.  The one exception is when  $R^2 = 0,$ where any increase in LLM sample size does not decrease the required number of human reviews, since the LLM does not provide any predictive ability. 

Consider an investigator with a target effect sample size of $n^* = 200$ and a human annotation budget of 100. If prior data between LLM and human ratings yields an estimate of $R^2 = 0.70,$ the investigator can work backward to select an LLM sample size $N$ based on budget constraints on the number of human reviews. With an LLM sample size of $N=2000$, only $65$ human samples are needed. If the budget can allow for 100 human ratings, then the LLM sample size $N$ can be reduced to $400$ while still achieving the target power.

\begin{figure}[htbp]
\centering
\includegraphics[width=0.8\textwidth]{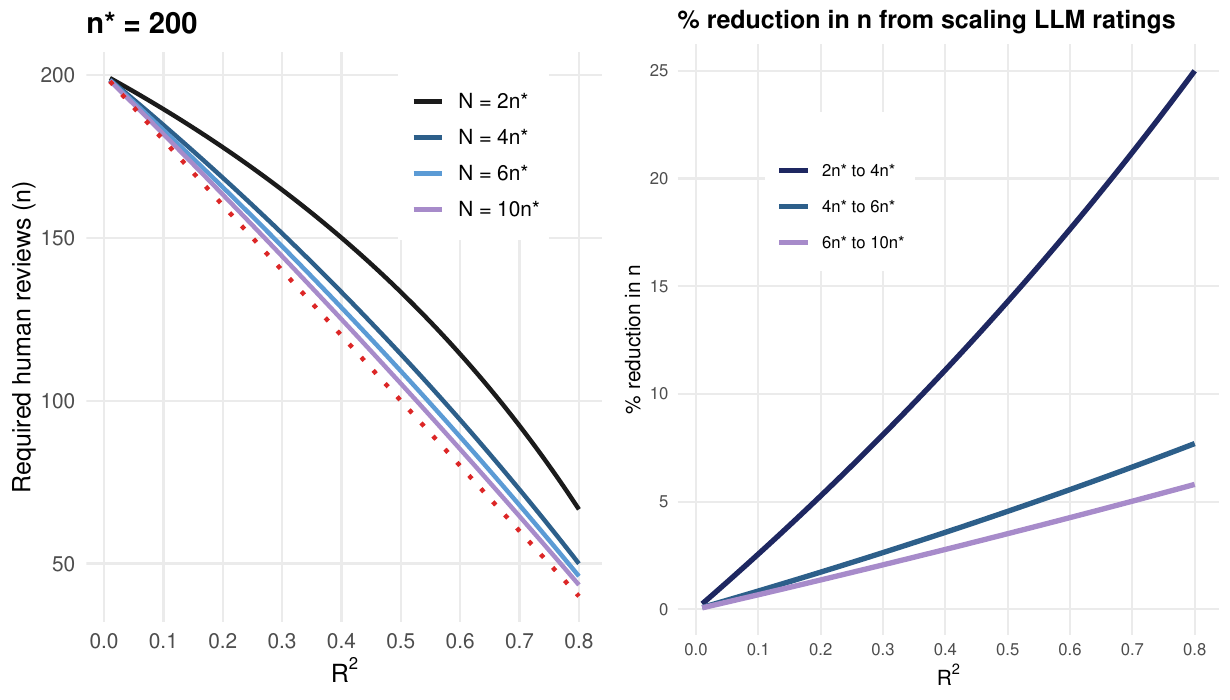}
\caption{(Left): Number of human reviews as a function of $R^2$, for a given target effective sample size. An increase in LLM reviews reduces human reviews for larger values of $R^2$. The minimum number of human samples for a given $R^2$, as $N$ increases to $\infty,$ is given by the red dotted line. (Right): The percent reduction in the required number of human samples increases as $R^2$ increases, but the marginal reduction decreases as the LLM ratings scale. }
\label{fig:Fig3}
\end{figure}

 \paragraph{Behavior of sample size.} Prediction quality $R^2$ determines the required number of human reviews, compared to the total LLM sample size $N.$ This follows from the property of the sample size formula. As $N$ increases, the required number of human reviews $n$ converges to a floor of $n^*(1-\rho^2)$. Two implications follow. One is that the benefit of increasing $N$ depends on $R^2.$ For $n^*=100$, when $R^2 = 0.1$, doubling the LLM pool from $N=200$ to $N=400$ reduces the required human reviews from 95 to 92, a negligible gain. When $R^2 = 0.8$, the same doubling reduces the requirement from 33 to 25, a 25$\%$ reduction. Second, each successive increase in $N$ produces diminishing reductions in $n$ (Figure~\ref{fig:Fig3}).  At $n^* = 200$ and $R^2 = 0.7$, a two-fold increase in LLM evaluations reduces the human budget by 20$\%$, though this reduction shrinks with each successive increase.

\subsection{Stratified sampling}

Now we can consider the case where response probabilities $\pi_i$ are different over response units $i$'s. That is, in the second stage, instead of using single sampling probability applied for the whole sample, one can sample units for human review through stratified sampling. The motivation comes from \cite{kumar2026} who demonstrated the heterogeneity of expert-LLM agreement across various evaluation dimensions of empathic communication. When $R^2$ varies across strata, it is natural to allow $\pi$ to be different. The strata for the case of \cite{kumar2026} would be various dimensions of empathic communication.

Consider the case where there are two strata $s = 1, 2.$ Let $\pi_i$ be $p_1$ if the $i^{th}$ unit belongs to $s=1,$ $p_2$ if $s=2,$ and $R_i^2$ will be $r_1^2$ if $s=1,$ and $r_2^2,$ if $s=2.$ The design task is to find the stratum-specific human evaluation sample sizes $n_1$ and $n_2$ respectively, given the stratum-specific LLM sample sizes $N_1, N_2,$ and stratum-specific correlations $ r_1^2, r_2^2.$

In this two-strata case, the variance formula becomes:

\begin{align}
 \frac{\sigma^2}{n^*}  & \approx  \frac{1}{N} \left[ \sigma^2 +  \frac{N_1}{N}   \frac{1 - p_1}{ p_1} \cdot \sigma^2(1 - r_1^2) +   \frac{N_2}{N}   \frac{1 - p_2}{ p_2} \cdot \sigma^2(1 - r_2^2) \right].  
\end{align}

\begin{figure}[ht]
\centering
\includegraphics[width=1.0\textwidth]{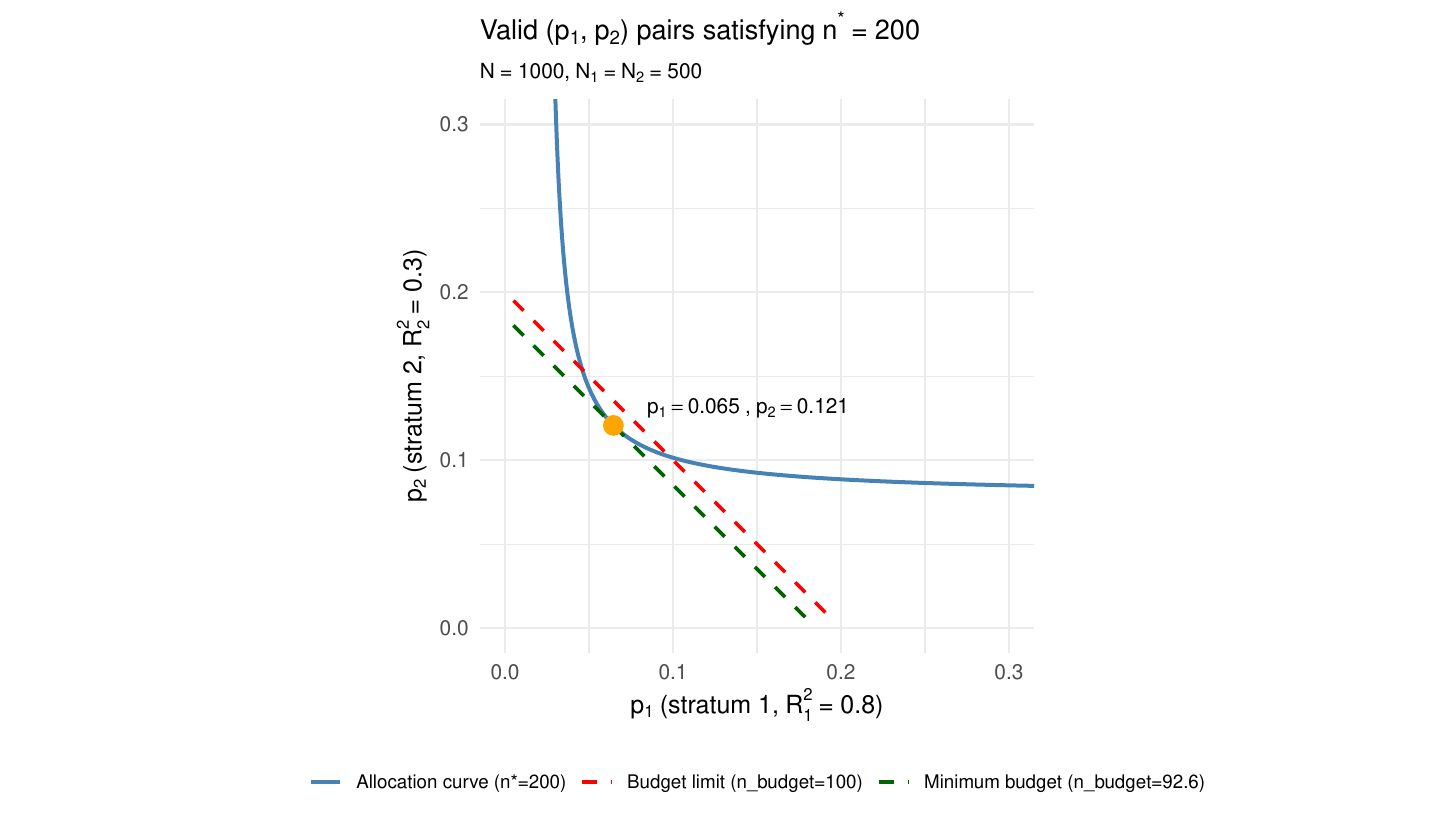}
\caption{Valid $(p_1, p_2)$ pairs satisfying the variance equation at $n^*=200,$ for $R_1^2=0.8$, $R_2^2=0.3$, $N=1000$. The budget line at $n_{\text{budget}}=100$ reflects the maximum budget allowed by the investigator. The minimum-cost design is shown by the green line.}
\label{fig:Fig4}
\end{figure}

\noindent Multiple numbers of pairs $(p_1, p_2)$ and thus $(n_1, n_2)$ satisfies the equation (4) given $(N_1, N_2, r_1^2, r_2^2),$ providing a valid design equivalent to achieving at least $n^*$. Figure~\ref{fig:Fig4} gives an example of an allocation curve, which shows that there is a minimum number of human ratings with a given setup.  If the human review budget stays above the minimum design, the investigator can choose the optimal number.  If the budget falls below, an increase in $(N_1, N_2)$ is necessary since $(r_1^2, r_2^2)$ is usually given, which will generate a new allocation curve.

Figure~\ref{fig:Fig4} shows the allocation curve, the set of sampling probabilities $(p_1, p_2)$ that achieve an equivalent power to $n^*=200,$ given $N_1=N_2=500,$ $r_1^2 = 0.8$ and $r_2^2 = 0.3.$ Any point on this curve is a valid design. The blue line, the allocation curve, shows the pairs of $(p_1, p_2)$ satisfying $n^*=200.$  The  red dotted line shows the budget constraint of  $500 p_1 + 500 p_2 = 100$. 
  The green dotted line shows the minimum cost design, where the budget line is tangent to the allocation curve at $p_1 = 0.065$ and $p_2 = 0.121,$ requiring at least $n_1 = 33$ and $n_2 = 61,$ for a total of 94 human ratings. By contrast, the design at   $p_1 =0.3$ and $p_2 = 0.09$ also satisfies $n^*=200,$ but requires $n_1 = 150$ and $n_2 = 45$, a total $n$ of 195, more than double the minimal budget. The difference arises because stratum 1 has a high predictive power $r_2^2=0.8$, and the minimum cost design exploits this by allocating fewer human reviews where the LLM is most reliable. 

\begin{figure}[htbp]
\centering
\includegraphics[width=0.75\textwidth]{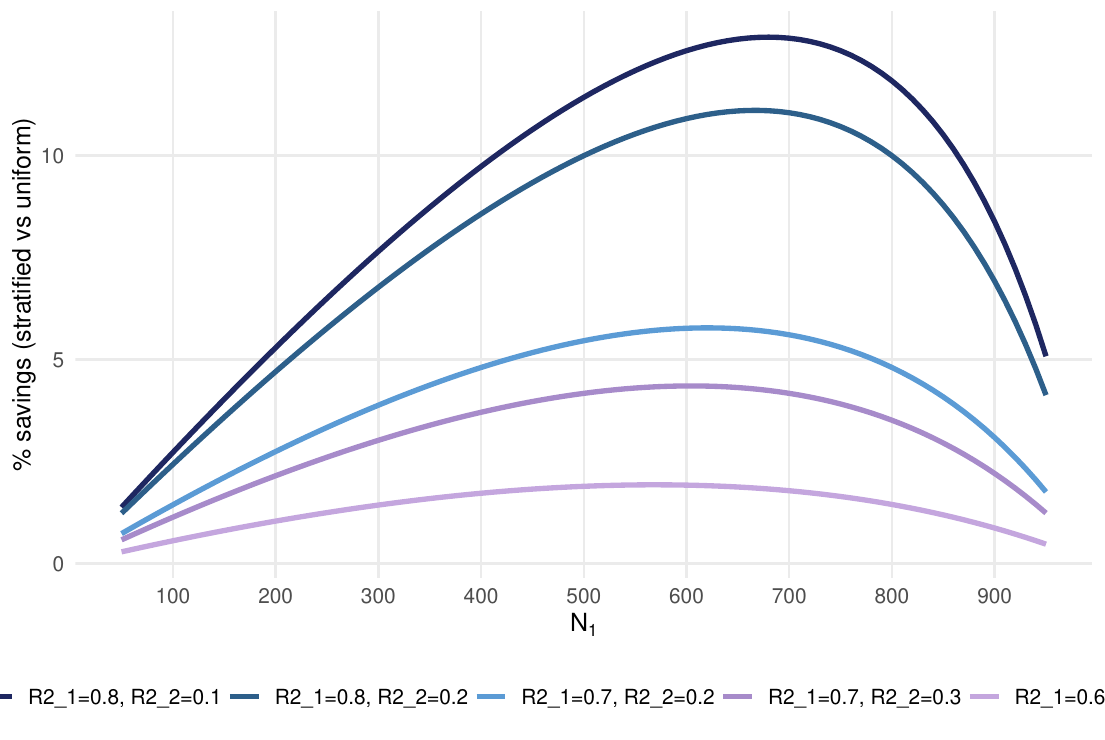}
\caption{Reduction in total human samples when comparing two-stratum vs. uniform sampling. The single-stratum design assumes a uniform LLM-human correlation of $R^2$ across all items, with a single sampling rate $\pi = n_{\text{budget}}/N$. The two-stratum design assumes heterogeneous LLM reliability, with $R_1^2$ in stratum 1 and $R_2^2$ in stratum 2. Both designs assume $N = 1000$ and ratios of $N_1$ and $N_2$ are varied.}
\label{fig:Fig5}
\end{figure}

\paragraph{Comparison of designs.}
We next compare stratified allocation where each stratum receives a different sampling probability, to uniform allocation where a single sampling probability is applied across all strata. Using a two-stratum design with stratum-specific $R^2$ values, we examine the reduction in total human reviews achieved by allowing the sampling probabilities to differ. Figure~\ref{fig:Fig5} shows the savings from stratification across a range of stratum configurations.

Stratified sampling yields the greatest savings when two conditions hold. First, the LLM predicts human ratings well on some dimensions but poorly on others, such that the difference in $R^2$ is large.  When the gap is large ($R_1^2 = 0.8$ and $R_2^2 = 0.1$), stratification reduces the human budget by up to 12.9$\%$ relative to uniform allocation. When the gap is smaller but moderate ($R_1^2 = 0.6$ and $R_2^2 = 0.3$) the savings barely exceed 2$\%$. Savings are also greater when the stratum where the LLM is most predictive accounts for a large proportion of the total evaluations. Under uniform allocation, this stratum receives a correspondingly large share of human reviews despite contributing little additional precision, and stratification reclaims that wasted effort.

\section{Discussion}\label{sec3}

This paper reformulates LLM evaluation as a statistical estimation problem in which human ratings are the primary quantity of inferential interest and LLM-generated ratings serve as auxiliary measurements. At present, no widely adopted methodology exists for determining how many items require human review.  

\paragraph{Related Work} 
Two prior works that are most closely related to our work examine  allocation strategies under fixed annotation budgets, though neither of these propose sample size formulas. \cite{mozer2026} derives a stratified model-assisted estimator and variance formula. The approach taken by \cite{mozer2026} is to take the sample size as given and then allocate the given budget, focusing on variance reduction from stratification. 

Our work addresses the prior step in the design process, which is to determine how large the human coding budget must be to achieve a target effective sample size. In our framework, our validity comes from the design that the investigator controls the response probabilities.  \cite{unell2025} derives a Chernoff bound for the sample size required to estimate the intraclass correlation coefficient between human and LLM judges within a specified tolerance. They also show that cluster-based selection of items for human annotation reduces ICC estimation error.

Another related but distinct area of work is active learning (\cite{settles2008}), which similarly recognizes that annotation is expensive and seeks to sequentially label instances based on current model uncertainty. Our approach differs in that human labels are the quantity of inferential interest, whereas in active learning human labels are used to train a model and improve predictive performance on future instances. 

\paragraph{Limitations} The sample size formula requires an $R^2$ value from a pilot study. If the pilot overestimates $R^2$ due to small sample size, overfitting, or a pilot corpus that is not representative of the full evaluation, the resulting design will underallocate human reviews. The effective sample size achieved will be lower than the target $n^*$ and the precision guarantees will not hold. This risk is not unique to this framework. Conventional sample size calculations face the same sensitivity to variance estimates from pilot data.  Investigators can guard against this by computing $n$ at a conservatively low $R^2$ or by conducting a sensitivity analysis across a plausible range.

This work assumes that human ratings constitute the gold standard against which LLM predictions are evaluated. This assumption is reasonable in settings where the construct being measured is well defined and disagreement among human raters reflects measurement noise rather than ambiguity in the construct itself. In many clinical evaluation contexts, however, the reliability of human ratings is itself uncertain, and the designation of human judgment as the gold standard is a pragmatic choice rather than a settled fact. The framework does not address the problem of unreliable or inconsistent human raters. 

\paragraph{Implications}

An implication of this framework is that high agreement between human and LLM ratings is not a prerequisite for the LLM to be useful in evaluation. Even when $R^2$ is modest, the LLM reduces the required number of human reviews relative to full human evaluation, and the sample size formula quantifies the magnitude of this reduction for any level of predictive quality. The framework directs human effort to the strata where it is most needed, rather than requiring that all strata meet a uniform threshold of agreement. Efforts to improve human-LLM agreement through prompt engineering or model selection, while potentially valuable, are not well motivated without reference to a target $n^*$ derived from the sample size calculation. 

As regulatory agencies develop standards for AI evaluation in healthcare, it is likely they will look to evidentiary standards that are not necessarily identical, but resemble gold standard practices in clinical studies. Agreement metrics will likely not suffice for this, but the specification of sample sizes, evaluation contexts, and target precision provide a path forward in closing the evidence and policy gap. The variance expressions and sample size formulas derived here provide a basis for such standards, analogous to the power calculations required for clinical trial protocols.  The pace of model development in clinical AI may seemingly provide fertile ground for relaxing evidentiary standards, but we argue that rapid pace of this evolving field demands principled validation design.  Researchers who must continually monitor and re-evaluate evolving models will benefit most from prospective designs that allocate human review efficiently.

\paragraph{Future Work} The case we described refers to one human rater. Future work can explore extending this case to multiple human raters. The examples covered in this paper are concerned primarily with discrete, single-turn evaluative tasks, and extending it to multi-turn dialogues and sequential clinical interactions will require additional considerations. 

\paragraph{Conclusions} 

%We proposed a shift in the formulation of the LLM evaluation problem: LLM ratings as auxiliary variables in an estimation problem with incomplete human ratings, which unifies several problems that the field has treated as separate. Sample size determination, allocation of human oversight, and the combination of human and LLM ratings are all governed by a single variance expression linking estimator precision to the degree of human-LLM agreement and the proportion of items reviewed. None of these quantities are available from the agreement metrics that currently dominate LLM evaluation practice. 

The framework presented in this paper provides the first principled basis for determining how many human labels are required to achieve a target level of inferential precision when LLM-generated ratings are used as auxiliary measurements through the principled combination of human and LLM ratings. It applies wherever the evaluation target is human expert judgment and annotation is scarce. Substituting LLM ratings for human judgment may be appropriate in low-stakes settings where the construct is well-defined and the consequences of error are limited. In domains where the constructs being evaluated involve normative judgment, such as safety assessment and mental health evaluation, expert ratings remain the standard against which LLM ratings are assessed,  and the need for principled human oversight is most acute.

%While replacing human ratings with LLM-generated ratings may be appropriate in low-stakes settings where the construct is well-defined and the consequences of error are limited, this substitution may not defensible as a first step in domains where expert judgment is consequential, such as safety assessment and mental health evaluation. In these settings, the constructs being evaluated are inherently normative, and LLM ratings are only meaningful as approximations of what a domain expert would have assigned. 

%The methods presented provide design guidelines for the LLM evaluation setting and are domain-agnostic. Any domain where LLMs are used to evaluate specialized content faces the same inferential problem: limited expert availability, high evaluation volume and unknown LLM reliability across contexts. The formulation presented applies wherever the evaluation target is human expert judgment and the LLM serves as an auxiliary measurement. This framework enables more rigorous, transparent evaluation of AI systems in high-stakes domains such as clinical AI. 

%\begin{ack}

%Author reports funding from a P50 grant from NIMH.  No Competing interests to report. 

%\end{ack}

\newpage

\bibliographystyle{plainnat}
\bibliography{references}

\end{document}